# Negative binomial regression and inference using a pre-trained transformer

Valentine Svensson [1]

[1] Tahoe Therapeutics, valentine@nxn.se

## Abstract

**Negative binomial regression is essential for analyzing over-dispersed count data in in comparative studies, but parameter estimation becomes computationally challenging in large screens requiring millions of comparisons. We investigate using a pre-trained transformer to produce estimates of negative binomial regression parameters from observed count data, trained through synthetic data generation to learn to invert the process of generating counts from parameters.**

**The transformer method achieved better parameter accuracy than maximum likelihood optimization while being 20 times faster.**

**However, comparisons unexpectedly revealed that method of moment estimates performed as well as maximum likelihood optimization in accuracy, while being 1,000 times faster and producing better-calibrated and more powerful tests, making it the most efficient solution for this application.**

## Introduction

In many situations where researchers perform comparative studies they observe discrete counts. For example, in genomics, copies of molecules are counted between treated and control samples. In clinical studies, occurrences of side effects are counted in treated and untreated patients. And in ecological studies, counts of numbers of species may be compared between two types of habitats.

When magnitudes of counts are large, they can be treated as normally distributed data after log transformation[1], but when magnitudes of counts are small this strategy fails[2]. The discrete nature of the data must be explicitly modeled.

In ideal situations, discrete counts can be compared through Fisher exact tests, hypergeometric tests, Binomial tests, or Poisson tests. These tests accounts for the variation in observed counts from the fundamental process of counting. However, variation between observations of counts often exceed what is expected by the count distributions assumed for these tests. There is variation in the data in addition to the process of counting. For example, in RNA-sequencing data, biological variability between replicates typically exceed Poisson assumptions, or, in epidemiological studies, patient heterogeneity can cause increased variation.

The solution to this problem is the application of count distributions with *over-dispersion*. A commonly used over-dispersed count distribution is the *Poisson gamma mixture* distribution, often referred to as the *negative binomial* distribution.

For individual comparative studies, computational speed when working with negative binomial distributions for regression analysis is negligible. However, increasing numbers of large scale screens are being produced. It is now feasible to create genome-wide screens which reads out molecule counts from every gene at once[3] comparing each gene knockout to a negative control would result in approximately $20,000 \cdot 20,000 = 4 \cdot 10^8$ binary comparisons. In such situations, accuracy of parameter estimates need to be considered in relation to the computational cost of acquiring the estimates.

Here, we create a novel method to estimate the parameters of negative binomial regression problems with a single binary predictor using a pre-trained transformer. We evaluate the transformer-based method and two classical parameter estimation methods for the regression problem.

**Negative binomial regression**

The negative binomial regression problem can be written as

$$y_i \sim \mathrm{NB}(l_i \cdot \exp(\mu + x_i \cdot \beta), \varphi),$$

where $x_i \in \{0, 1\}$. Given the observed data $\{x_i, y_i, l_i\}$ the problem is to estimate the parameters $\theta = \{\mu, \beta, \varphi\}$.

In this formulation, $\mu$ is the base mean and $\beta$ is the effect size if there is a difference between the groups of obser-vations indexed by $x$. The parameter $\varphi$ represents the amount of over-dispersion, how much more variation there is between observations of counts than one would expect when counting occurrences.

The dispersion parameter $\varphi$ is challenging to fit accurately, partially due to a lack of identifiably in the negative binomial regression model. At the same time, accurate estimation of the $\varphi$ parameter is crucial for statistical inference as $\text{SE}(\hat{\beta})$ directly depends on $\varphi$.

**Parameter estimation through maximum likelihood optimization**

The Python package statsmodels[4] uses Newton-Raphson optimization to maximize the log likelihood $\ell$,

$$\hat{\theta} = \arg\max_{\theta} \ell(\theta),$$

using analytical gradients

$$\frac{\partial \ell}{\partial \beta} = \sum_i x_i \cdot \left[ \psi(y_i + a) - \psi(a) + \log(r_i) - \frac{(y_i + a) \cdot m_i}{a + m_i} \right]$$

(and equivalent for $\mu$), and

$$\frac{\partial \ell}{\partial \varphi} = \sum_i \left[ \frac{1}{\varphi^2} \cdot \left( \psi(a) - \psi(y_i + a) - \log(r_i) + \frac{(y_i + a) \cdot m_i}{a + m_i} \right) \right],$$

where $\psi(\cdot)$ is the digamma function. The implementation uses an internal parameterization where

$$a = \frac{1}{\varphi},$$
$$m_i = l_i \cdot \exp(\mu + x_i \cdot \beta),$$
$$r_i = \frac{a}{a + m_i}.$$

Maximum likelihood optimization is flexible, and can be applied to much more complex experimental designs than considered here. It is theoretically optimal, achieving the the lowest possible variance among unbiases estimators. On the other hand, iterative numerical optimization is computationally expensive and requires handling initialization[5].

Maximum likelihood optimization for the negative binomial regression problem is implemented in the class sm.NegativeBinomial[1].

---
[1]Note that the class sm.GLM with family = sm.families.NegativeBinomial() in statsmodels does not implement full negative binomial regression, because it assumes the dispersion parameter $\varphi$ is known and fixed.

**Parameter estimation through method of moments**

As a fast, non-iterative, alternative to maximum likelihood optimization the method of moments can be used to estimate the parameters $\theta$. If $Y_1 = \{y_i \mid x_i = 0\}$ and $Y_2 = \{y_i \mid x_i = 1\}$ (and similarly for $L_1$ and $L_2$), then the method of moment estimates for the parameters $\theta$ are

$$\hat{\mu} = \log\left(\frac{\overline{Y_1}}{\overline{L_1}}\right), \ \hat{\beta} = \log\left(\frac{\overline{Y_2}}{\overline{L_2}}\right) - \hat{\mu}, \ \hat{\varphi} = \frac{S^2 - \overline{Y}}{\overline{Y_2}},$$

where $\overline{Y}$ represents the arithmetic mean of the elements of the set $Y$.

While method of moments is extremely fast for estimation, it typically requires larger samples sizes than maximum likelihood optimization to achieve similar performance (10-30% more observations)[5].

**Parameter estimation with a pre-trained transformer**

A set transformer[6] function $f_\varphi$ was trained to map sets of observations to generative parameters:

$$f_\varphi : (Y_1, L_1, Y_2, L_2) \mapsto \hat{\theta} = \{\hat{\mu}, \hat{\beta}, \hat{\varphi}\}.$$

Observations are transformed through $y_i^\star = \log_{10}(10^4 \cdot y_i / l_i + 1)$ before being passed to the transformer stem. The transformer is designed with self-attention within the sets $Y_1^\star$ and $Y_2^\star$ to learn to interpret within-group variation, and cross-attention between the sets $Y_1^\star$ and $Y_2^\star$ to learn to interpret how between-group variation relates to the within-group variation.

Similar to method of moments, this is a non-iterative estimation method, though with far more complex computations.

The transformer was trained using synthetic data generation of counts simulated from negative binomial distributions with known parameters. In this way, the transformer learns to invert the process of generating count observations.

## Results

Our primary goal is to investigate which methods have sufficient accuracy in parameter estimation with the least computational burden. We evaluated this by simulating 10,000 regression problems with three control observations and three treatment observations. All methods can be used with varying numbers of replicates, but in many situations the number of replicates are fixed for all comparisons, enabling fast vectorized operations.

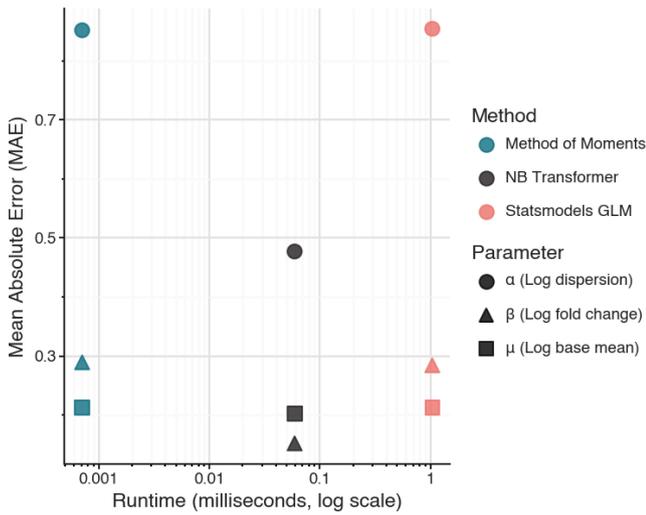

Figure 1: Average runtime vs accuracy across 10,000 simulated estimation problems. All methods evaluated on the same CPU using vectorization for runtime performance

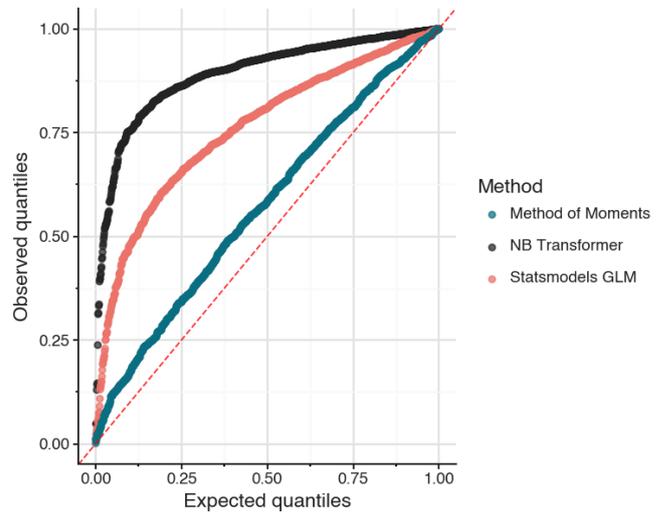

Figure 2: QQ-plot comparing calibration of the significance test with null hypothesis $\beta = 0$, based on 1,000 simulations.

All methods produce more accurate estimates for $\mu$ and $\beta$ than for dispersion $\varphi$ (Figure 1). Surprisingly, method of moments estimation is as accurate as maximum likelihood optimization even though it is commonly thought that maximum likelihood is more sample efficient[5].

The transformer method yield marginally more accurate estimates for $\mu$ and $\beta$, but substantially more accurate estimates for $\varphi = \exp(\alpha)$.

The iterative maximum likelihood optimization estimation requires most computational time, while the simple method of moments estimation is the fastest, being 1000 times faster. The transformer method is 20 times faster than maximum likelihood optimization[2].

Following parameter estimation, a researcher typically want to assess how strong the evidence is for an estimated effect. This is commonly done through null hypothesis significance testing, obtaining a p-value for the null hypothesis $\beta = 0$. For a properly calibrated significance test the distribution of p-values is uniform when the null hypothesis is true.

To assess calibration, we simulate 1,000 experiments with $\beta = 0$ and investigate the QQ-plot of the distributions of analytically derived p-values.

Surprisingly, the method of moments estimates lead to the best calibrated p-values. While not optimal, method of moments have only slightly conservative p-values, meaning erroneously accepting the null hypothesis (Figure 2).

Both maximum likelihood optimization and the transformer method leads to substantially conservative p-values, with the transformer method being the most over-conservative.

Finally, we want to compare the *power* of the statistical tests from the different parameter estimation methods. That is, when there is a true effect, how likely are we to correctly reject the null hypothesis, depending on the size of the true effect?

For 10 different known beta values ranging between 0 and 2.5, we sample 1,000 experiments with four different replication designs (3v3, 5v5, 7v7, and 9v9). Each of the 40,000 simulated experiments are used to perform parameter estimation with the three different methods, and the parameters are used for analytical p-value calculation. Power is calculated as the fraction of tests were the null hypothesis is correctly rejected by a threshold of $p < 0.05$.

---

[2]All runtime comparisons were executed on CPU using an Apple M4 Pro.

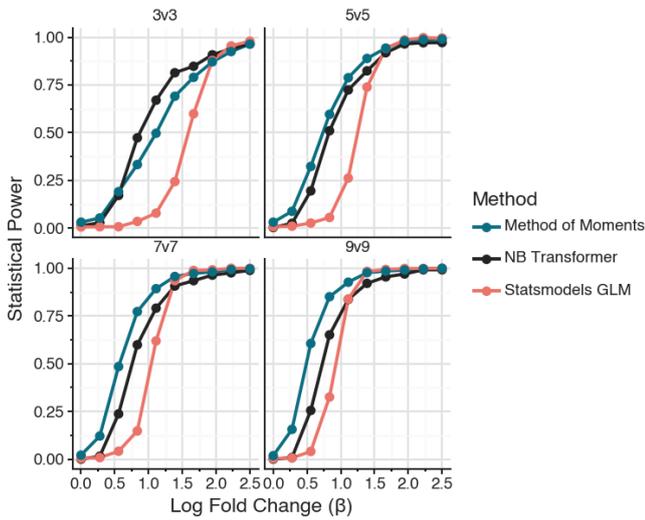

Figure 3: Power analysis of the parameter estimation using different known effect sizes and experimental designs. For each power calculation, 1,000 experiments were simulated.

Maximum likelihood optimization estimates have the lowest power, while estimates from method of moments and the transformer model has roughly equal power. The transformer model has marginally higher power at the smallest sample size (3v3). Interestingly, while power increases with sample size, the increase in power over sample sizes is more marginal for the transformer model. At the largest sample size (9v9) the method of moment estimates have the highest power (Figure 3).

## Discussion

As technological development advances, it is getting increasingly feasible to collect large amounts of data. The most intuitive and direct analysis of data is a comparison between a control group and a group of interest.

Here, we have introduced a novel transformer-based method to estimate parameters of a comparative model of integer counts with the aim of increasing estimation speed.

The transformer model is trained on synthetically generated data with the goal of inverting the stochastic data generation process. It is able to do this by learning within-set and between-set variabilities, and how they relate to the true generative parameters.

We compared accuracy and run times of the transformer model with two classical parameter estimation approaches: maximum likelihood optimization and method of moments. Overall, the transformer had slightly increased accuracy performance than the classical methods. It had 20 times faster runtime than maximum likelihood optimization, but method of moments had 1000 times faster runtime.

Surprisingly, method of moment estimates were as accurate as maximum likelihood optimization, even though method of moments is typically assumed to have poor performance with low sample sizes.

Maximum likelihood optimization performs poorly in runtime, accuracy, and statistical properties. On the other hand, maximum likelihood optimization is very flexible, and can be applied with arbitrary design matrices. Method of moments estimators need to be derived for particular questions with special experimental designs, and transformer models need to be designed and trained for new questions.

While the novel direction of using a pre-trained transformer for statistical estimation is promising, for this particular problem application of method of moments would be the most efficient solution.

In the field of gene expression analysis, specialized tools have been developed for testing differences in molecule counts of genes between treatment and control, such as DESeq2[7]. These statistical methods differ from the methods presented here in that they are hierarchical models that pool information between genes, performing shrinkage of the dispersion parameter $\varphi$ and effect size $\beta$ towards averages across genes.

A potentially interesting future direction could be to include cross-attention between multiple sets of observations which could be used to learn group-level parameters in hierarchical models with a transformer-based model.

An implementation of the transformer model with pre-trained weights is available on Hugging Face: https://huggingface.co/valsv/nb-transformer, the repository also contains scripts for producing the analyses presented here.

## Methods

### Deriving method of moments estimators

The first moment (mean) of the distribution is

$$E[y_i] = m_i = l_i \cdot \exp(\mu + \beta \cdot x_i),$$

and the second moment (variance) of the distribution is

$$\operatorname{Var}(y_i) = m_i + \varphi \cdot m_i^2.$$

For condition 1 data ($x_i = 0$), we have

$$E[y_i \mid x_i = 0] = l_i \exp(\mu).$$

By taking the expectation over exposures $l_i$ we get

$$E[y_i \mid x_i = 0] = E[l_i] \exp(\mu)$$

and can solve for the method of moment estimator

$$\hat{\mu} = \log\left(\frac{\overline{Y_1}}{\overline{L_1}}\right).$$

For condition 2 data ($x_i = 1$) we have

$$E[Y_i \mid x_i = 1] = l_i \cdot \exp(\mu + \beta) = l_i \cdot \exp(\mu) \cdot \exp(\beta).$$

The ratio of the conditional means gives us

$$\exp(\beta) = \frac{E[Y_i \mid x_i = 1] / E[l_i \mid x_i = 1]}{E[Y_i \mid x_i = 0] / E[l_i \mid x_i = 0]},$$

which gives us the methods of moment estimator

$$\hat{\beta} = \log\left(\frac{\overline{Y_2}/\overline{L_2}}{\overline{Y_1}/\overline{L_1}}\right) = \log\left(\frac{\overline{Y_2}}{\overline{L_2}}\right) - \hat{\mu}.$$

Since

$$\mathrm{Var}(y) = E[y] + \varphi \cdot E[y]^2,$$

we have

$$\varphi = \frac{\mathrm{Var}(y) - E[y]}{E[y]^2},$$

from which we get the method of moments estimator

$$\hat{\varphi} = \frac{S^2 - \overline{Y}}{\overline{Y}^2}.$$

**Analytic expression for effect size standard error**

The negative binomial regression model has design matrix $X = (\mathbf{1}\ x) \in \mathbb{R}^{n \times 2}$ where the predictor $x \in \{0, 1\}$. Each observation $i$ can be converted to a weight $w_i = \frac{m_i}{1+\varphi \cdot m_i}$. We can construct group-wise total weights through

$$S_0 = \sum_{x_i=0} w_i, \quad S_1 = \sum_{x_i=1} w_i.$$

Then the Fisher information matrix is the 2x2 matrix

$$X^T W X = \begin{pmatrix} S_0 + S_1 & S_1 \\ S_1 & S_1 \end{pmatrix},$$

with inverse

$$(X^T W X)^{-1} = \frac{1}{S_0 S_1} \cdot \begin{pmatrix} S_1 & -S_1 \\ -S_1 & S_0 + S_1 \end{pmatrix}.$$

The (2, 2) element of the inverse of the Fisher information matrix gives the variance of the estimate $\hat{\beta}$:

$$\mathrm{Var}(\hat{\beta}) = \frac{S_0 + S_1}{S_0 \cdot S_1} = \frac{1}{S_0} + \frac{1}{S_1}.$$

In other words, the standard error of effect size $\beta$ the can be directly calculated from the parameter estimates,

$$\mathrm{SE}(\hat{\beta}) = \sqrt{\frac{1}{\sum_{x_i=0} w_i} + \frac{1}{\sum_{x_i=1} w_i}}.$$

For significance testing, the Wald statistic $z = \frac{\hat{\beta}}{\mathrm{SE}(\hat{\beta})}$ can be converted to a two-sided p-value through either

$$p = 1 - F_{\chi^2(1)}(z^2)$$

or

$$p = 1 - F_{\mathrm{N}(0,1)}(z),$$

which yield equivalent results.

**Transformer architecture**

Counts $y_i$ and exposures $l_i$ are transformed to input values $y_i^\star = \log_{10}(10^4 \cdot y_i / l_i + 1)$. Each $y_i^\star$ is projected to a $d$-dimensional embedding. The transformer is designed for sets with at least two elements and at most 10 elements.

Within each set $Y_1^\star$ or $Y_2^\star$ the embeddings pass through $L$ multi-head self-attention blocks with $h$ heads followed by feed-forward sub-layers. The block learns exchangeable summaries of each set.

The representations of $Y_1^\star$ attent to those of $Y_2^\star$, and vice-versa. This allows each element to compare itself with the full distribution of the opposite condition and capture treatment-specific changes.

Mean-pooling produces two fixed-length vectors $\phi_1$ and $\phi_2$, one per set. These pooled representations are combined using bilinear interactions to capture comparative statistics:

$$\xi = [\phi_1; \phi_2; \phi_1 - \phi_2; \phi_1 \odot \phi_2] \in \mathbb{R}^{4 \cdot d}$$

The combined features $\xi$ are passed through a multi-layer perceptron head which produces predictions for the regression targets $\mu, \beta$ and $\alpha$, where $\alpha = \log(\varphi)$ (Figure 4).

Both inputs and targets are scaled by constant learned factors to make them closer to mean 0 and variance 1 which stabilizes training.

In the final trained model, $d = 128, h = 8, L = 3$ with a dropout rate of 0.1, for a total of 2.5 million parameters.

**Training through synthetic data generation**

We train the transformer model with a simple weighted loss

$$\mathcal{L}(\theta, \hat{\theta}) =$$
$$w_\mu \cdot (\mu - \hat{\mu})^2 + w_\beta \cdot (\beta - \hat{\beta})^2 + w_\alpha \cdot (\alpha - \hat{\alpha})^2,$$

with weights $w_\mu = 1.0, w_\beta = 1.0, w_\alpha = 2.0$. The higher weight for $\alpha$ emphasizes learning to predict the difficult-to-estimate but crucial dispersion parameters.

In each "epoch", we sample $N_{\mathrm{epoch}}$ parameter vectors $\theta^{(i)} \sim p(\theta)$ (including exposures $l^{(i)}$). For each $\theta^{(i)}$, we generate data $\left(Y_1^{(i)}, Y_2^{(i)}\right) \sim p\left(Y \mid \theta^{(i)}\right)$ with $N_1^{(i)}$ and $N_2^{(i)}$ observations per set. The sampled parameters and generated data are used as training pairs.

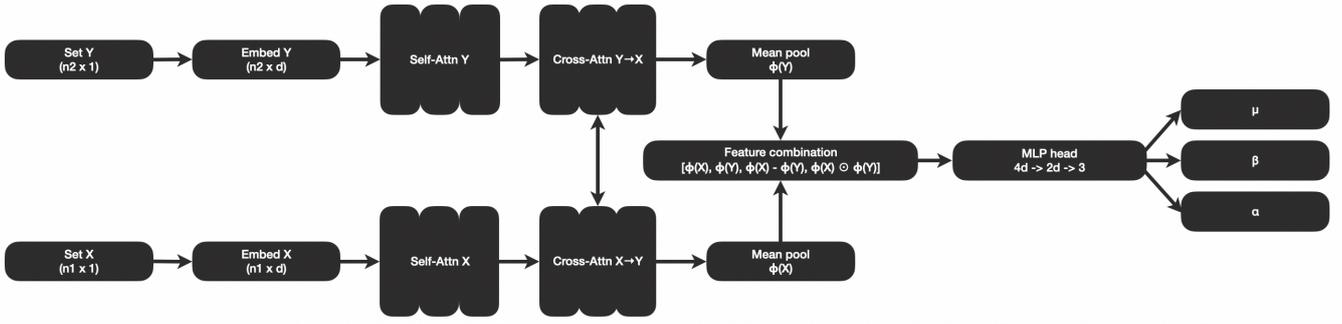

Figure 4: Architecture of the set transformer model for negative binomial regression.

Parameters are samples from the distributions

$$\mu \sim \mathrm{N}(-1, 2),$$
$$\alpha \sim \mathrm{N}(-2, 1),$$
$$\delta \sim \mathrm{Bernoulli}(0.3),$$
$$\beta \sim 0 + \delta \cdot \mathrm{N}(0, 1),$$
$$l \sim \mathrm{LogNormal}\left(\log(10^4) - \frac{\log(1.09)}{2}, \log(1.09)\right),$$
$$N_1 \sim \mathrm{U}_{\mathbb{Z}}(2, 10),$$
$$N_2 \sim \mathrm{U}_{\mathbb{Z}}(2, 10).$$

The same distributions are used for training and in the performance benchmarks.

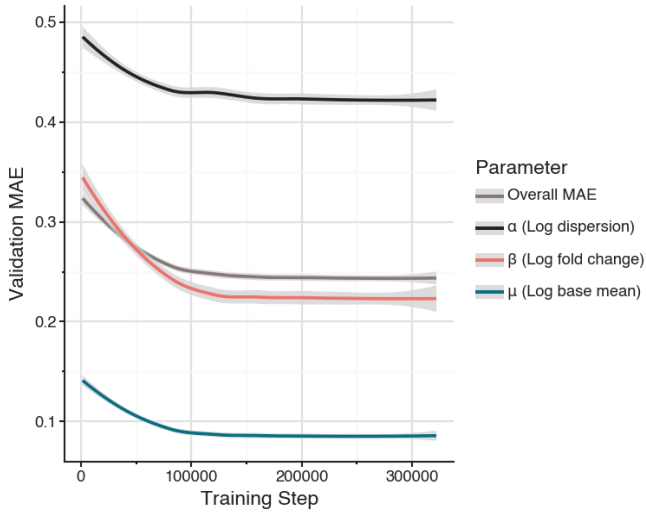

Figure 5: Validation error over training steps during training with synthetic data generation.

Optimization was performed using AdamW with `ReduceLROnPlateau` and a base learning rate of $10^{-4}$. Weight decay was set to $10^{-4}$. In training, $N_{\mathrm{epoch}}$ was set to 100,000 with a batch size of 32. The model was trained for 100 epochs using Apple MPS on an M4 Pro with 12 cores and 64GB unified memory (Figure 5).